\documentclass{article}

\usepackage[nonatbib, preprint]{neurips_data_2021}

\usepackage[utf8]{inputenc} % allow utf-8 input
\usepackage[T1]{fontenc}    % use 8-bit T1 fonts
\usepackage{hyperref}       % hyperlinks
\usepackage{url}            % simple URL typesetting
\usepackage{booktabs}       % professional-quality tables
\usepackage{amsfonts}       % blackboard math symbols
\usepackage{nicefrac}       % compact symbols for 1/2, etc.
\usepackage{microtype}      % microtypography
\usepackage{xcolor}         % colors
\usepackage{epsfig}
\usepackage{graphicx}
\usepackage{multirow}
\usepackage{booktabs}
\usepackage{caption} 
\usepackage{rotating}
\title{A Novel Dataset for Keypoint Detection of Quadruped Animals from Images}

\author{%
  Prianka Banik \\
  Department of Computer Science\\
  Prairie View A\&M University\\
  \texttt{pbanik@pvamu.edu} \\
   \And
  Lin Li \\
  Department of Computer Science\\
  Prairie View A\&M University\\
  \texttt{lilin@pvamu.edu} \\
   \And
  Xishuang Dong \\
  Department of Electrical and Computer Engineering\\
  Prairie View A\&M University\\
  \texttt{xidong@pvamu.edu} \\    
}

\begin{document}

\maketitle

\begin{abstract}
   In this paper, we studied the problem of localizing a generic set of keypoints across multiple quadruped or four-legged animal species from images. Due to the lack of large scale animal keypoint dataset with ground truth annotations, we developed a novel dataset, AwA Pose, for keypoint detection of quadruped animals from images. Our dataset contains significantly more keypoints per animal and has much more diverse animals than the existing datasets for animal keypoint detection. We benchmarked the dataset with a state-of-the-art deep learning model for different keypoint detection tasks, including both seen and unseen animal cases. Experimental results showed the effectiveness of the dataset. We believe that this dataset will help the computer vision community in the design and evaluation of improved models for the generalized quadruped animal keypoint detection problem.
\end{abstract}
\section{Introduction}
Detecting keypoints of quadruped animals has several practical applications in animal behavior understanding \cite{mathis2020deep}, automated identification and tracking \cite{t2020long}, part segmentation \cite{Naha_2021_WACV}, etc. Different quadruped animals vary highly in shapes and they can show wide range of poses. These make the keypoint detection of quadruped animal in wild extremely challenging compared to the intensively studied problem of human keypoint detection from images. 

Large scale visual data with manual annotations are required to train the state-of-the-art deep learning based keypoint detection models.  However, to the best of our knowledge, there are no large scale datasets of diverse quadruped animal images with corresponding keypoint annotations. Most existing datasets either focus on a single animal or only very few animals \cite{li2019atrw, graving2019fast, labuguen2020macaquepose, mathis2021pretraining,Cao_2019_ICCV} or focus on only a very small set of keypoints e.g. facial keypoints \cite{rashid2017interspecies}. Such datasets are inadequate to train and evaluate large deep learning based models for practical applications. Although quadruped animals have generalized body structures and common body parts, they often come with widely different sizes and shapes. For example, if a dataset contains only known quadruped animals such as dog, cat, horse, cow, sheep, etc. and corresponding keypoints, a model trained with such a dataset may fail to detect the keypoints of new animals like elephant, giraffe, ant-eater, rabbit, kangaroo, etc. Similarly, if a dataset contains large number of animals but only few keypoints such as the facial keypoints, a model trained with such a dataset will have limited usage to only face related applications. Full body related problems such as 3d model generation, action recognition, and motion analysis cannot be solved with such datasets. 

To address this problem, we built a large scale dataset for the quadruped animal keypoint detection problem. The dataset was named AwA Pose as the images were collected from the AwA dataset \cite{xian2018zero}. To increase the diversity of the images, we included a number of different species in our dataset. By doing so, we ensure that the dataset contains quadruped animals of different sizes and shapes, and therefore, a model trained on the dataset can be applied to unseen quadruped animals. Meanwhile, we also ensure that the animals in the images contains various poses as some animal classes (e.g., cat) can have many different poses which often make keypoint detection very challenging. The images in the dataset contain both self and other forms of occlusions which makes the dataset more generalized. In addition, we proposed a generic set of keypoints for all quadruped animals. The keypoint set includes almost all the body parts of the animals and thus makes the dataset useful for various practical applications. 

Besides developing the dataset, we benchmarked it using a state-of-the-art deep learning model for keypoint detection. Experimental results show that the model trained on the dataset can be successfully applied to both seen animals with different poses and novel animals with same and different poses. This ensures that the dataset contains enough diversities to train a generalized keypoint detection model.

Through this work, we intend to give the computer vision research community a new venue to develop and evaluate computer vision models for the quadruped animal keypoint detection problem. The dataset is available to public through github repository \url{https://github.com/prinik/AwA-Pose}.

\section{Related Works}

Keypoint detection of objects is a highly studied topic in computer vision. Most of the previous work focused on understanding human keypoints due to its applications in human computer interaction \cite{liu2021anisotropic}, action recognition \cite{holte2012human, luvizon20182d, yao2011does}, human pose transfer \cite{neverova2018dense, zanfir2018human}, 3D shape estimation \cite{kocabas2020vibe, omran2018neural}, etc. Compared with these efforts, very few had explored the problems of animal keypoint, part or shape estimation from images. Zuffi et. al. \cite{zuffi20173d} explored the problem of novel quadruped animal 3D shape estimation by utilizing 3D scans of toy figurines due to the lack of 3D scan data of real quadruped animals. Some previous research also explored estimating the keypoints of animals using self-supervision approaches \cite{zhang2018unsupervised, belikov2020goodpoint, thewlis2017unsupervised}, but these keypoints are difficult to evaluate as there are no corresponding ground truth data. Some previous work limited the problem to only facial keypoint detection from animal images \cite{rashid2017interspecies, yang2016human}. 
Recently, Cao et al \cite{cao2019cross} combined the keypoint annotation of human and quadruped animals to train a deep learning model which showed superior performance for the whole body keypoint estimation problem of novel quadruped animals. But they did not provide any quantitative results on novel quadruped animals due to the lack of existing large scale dataset of quadruped animals with keypoint annotations. Also, many animals have unique keypoints which are missing in human beings (e.g., tail end and horns). Moreover, their seen and unseen animals are pretty similar in shapes. It is not clear how well their approach will perform for animals with unusual body features (e.g., giraffe---long neck, elephant---trunk, etc.)

The need for a large scale quadruped animal keypoint dataset becomes more evident when the state-of-the-art deep learning models for keypoint detection such as HRNet \cite{sun2019deep} are used. Such models requires large number of annotated data for training. Also, without a proper testing dataset, it is difficult to understand the performance improvements of the new models. Existing datasets such as Animal Pose \cite{Cao_2019_ICCV} have few body keypoints for only five species of quadruped animals. The sparse keypoints are not very useful for problems such as 3D shape reconstructions. All these drove us to create a large scale quadruped animal keypoint dataset with diverse animals and a large number of keypoints.

\section{Our Dataset}
The goal of this paper is to introduce a new large scale dataset of quadruped animal keypoint---AwA Pose. We built the dataset using images from the AwA \cite{xian2018zero} dataset. AwA contains images of 50 animal species from which we selected 35 quadruped animal species. Compared with the existing animal keypoint datasets, AwA Pose better facilitates the training and testing of deep learning models with both more animals species and higher number of keypoints. For example, Animal Pose \cite{Cao_2019_ICCV} contains five animal species which is higher than the other datasets but still much lower than the number of animal species covered by our dataset. On the other hand, Horse-10 \cite{mathis2021pretraining} has the largest set of keypoints after ours, but it only contains a single animal `Horse' and the number of keypoints is much smaller than AwA Pose. The MacaquePose \cite{labuguen2020macaquepose} dataset contains the most number of images (~13k) where our dataset contains a comparable number of images (~10k) and much more animals and keypoints. A detailed comparison of the datasets is shown in Table \ref{table:datasets}. In the following sections, we will discuss in details about the keypoint set and the annotation process.  

\begin{table}
\centering
\begin{tabular}{llll} 
\hline
\textbf{Dataset} & \textbf{Images} & \textbf{Species} & \textbf{Keypoints}  \\ 
\hline
Animal Pose \cite{Cao_2019_ICCV}                                       & 3608      & 5       & 20          \\
Horse-10 \cite{mathis2021pretraining}                                          & 8114      & 1       & 22          \\
MacaquePose \cite{labuguen2020macaquepose}   & 13083      & 1       & 17          \\
Grévy’s Zebra \cite{graving2019fast} & 900      & 1       & 9          \\
ATRW \cite{li2019atrw}         & 8076      & 1       & 15          \\
\textbf{AwA Pose (Our Dataset)}                                          & \textbf{10064}      & \textbf{35}      & \textbf{39}          \\
\hline
\end{tabular}

\caption{List of selected generic keypoint datasets for the quadruped animals.}
\label{table:datasets}
\vspace{-6pt}
\end{table}

\subsection{Keypoint Set}
\begin{figure*}
\begin{center}
\includegraphics[width=14cm]{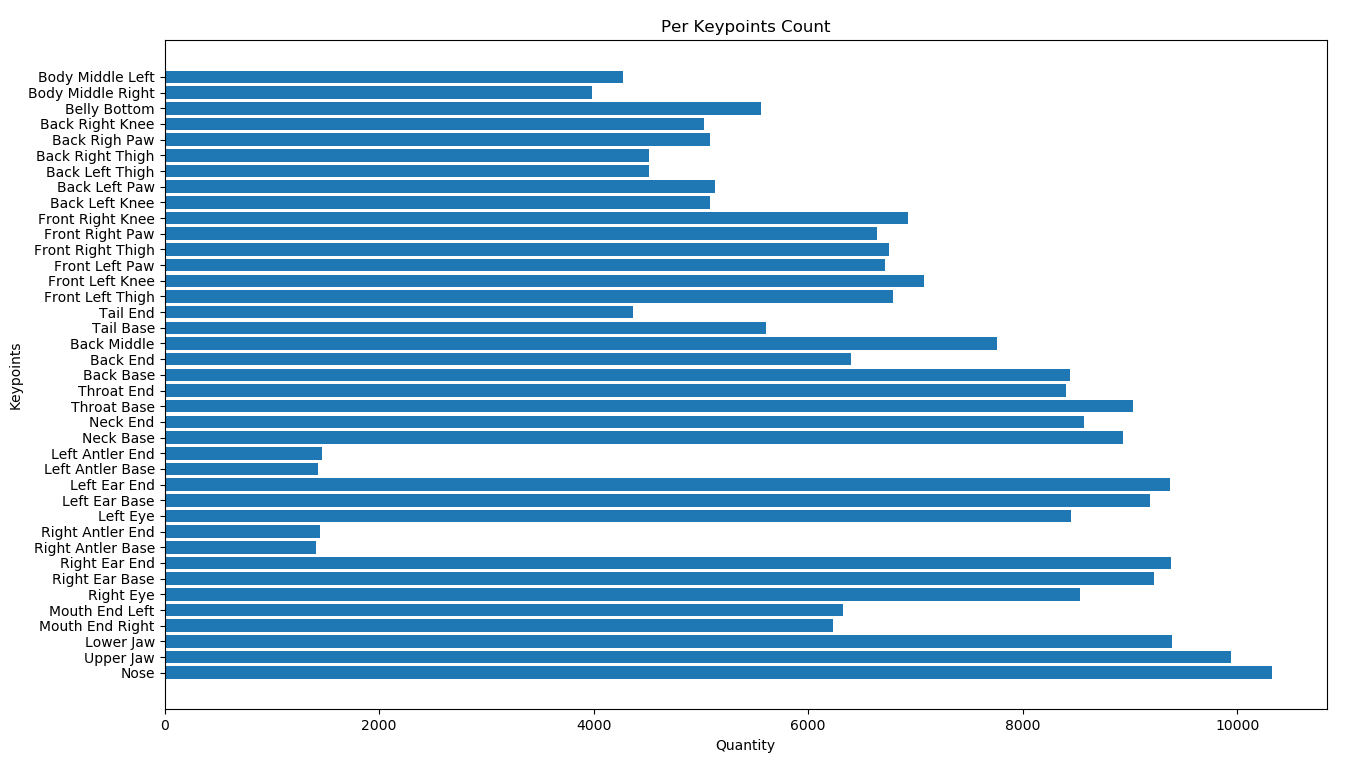}
\end{center}
  \caption{Number of images per keypoint.}
\label{fig:per_keypoint}
\end{figure*}

We first fixed a generic set of keypoints for all the selected quadruped animals. While most of the existing datasets only provide annotation for a very small number of keypoints, our dataset provides a much larger number of keypoints for better understanding the animal's poses. Our goal is to annotate as many keypoints as possible so we can cover almost all the body parts of the animal. This will help applications which requires dense keypoints such as 3D reconstruction, pose estimation, etc. The keypoints we adopted are mostly from Animal Pose \cite{Cao_2019_ICCV}. We augmented the keypoints by adding more details such as `Tail End', `Mouth End Left', `Mouth End Right', etc. A list of the 39 keypoints and their corresponding numbers of images is shown in Figure \ref{fig:per_keypoint}. It should be noted that we only annotated the visible keypoints in a specific picture. Figure \ref{fig:per_keypoint} shows that the upper body keypoints and facial keypoints are visible in most of the images. Keypoints related to the body part `Antler' are most rare as only few animals have `Antler'. Other than `Antler', all other keypoints have around 4k annotated images.

\subsection{Annotation}
\begin{figure*}
\begin{center}
\includegraphics[width=12cm]{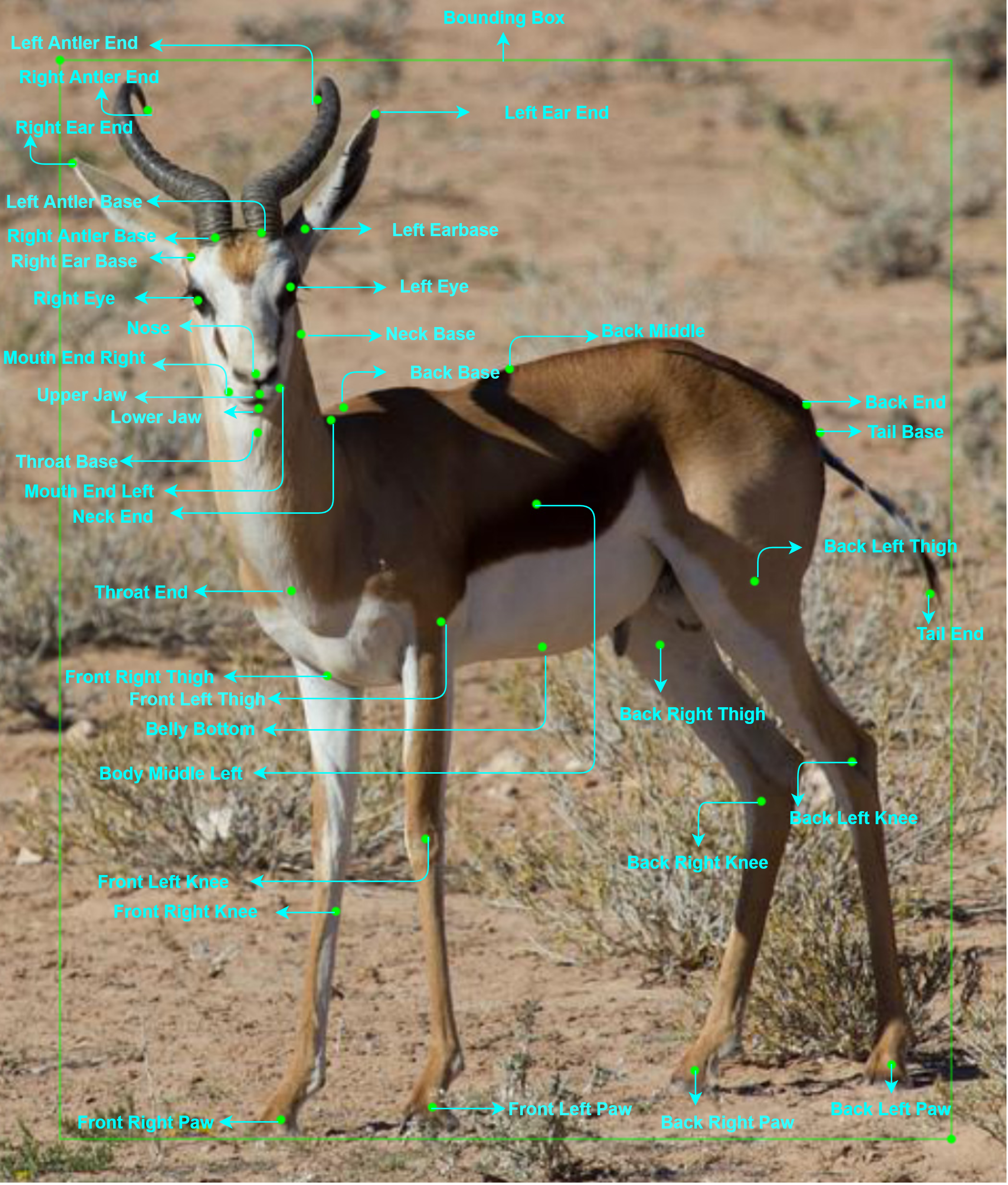}
\end{center}
  \caption{Example annotation of a quadruped animal with keypoints and bounding box.}
\label{fig:labelme}
\end{figure*}

\begin{figure*}
\begin{center}
\includegraphics[width=14cm]{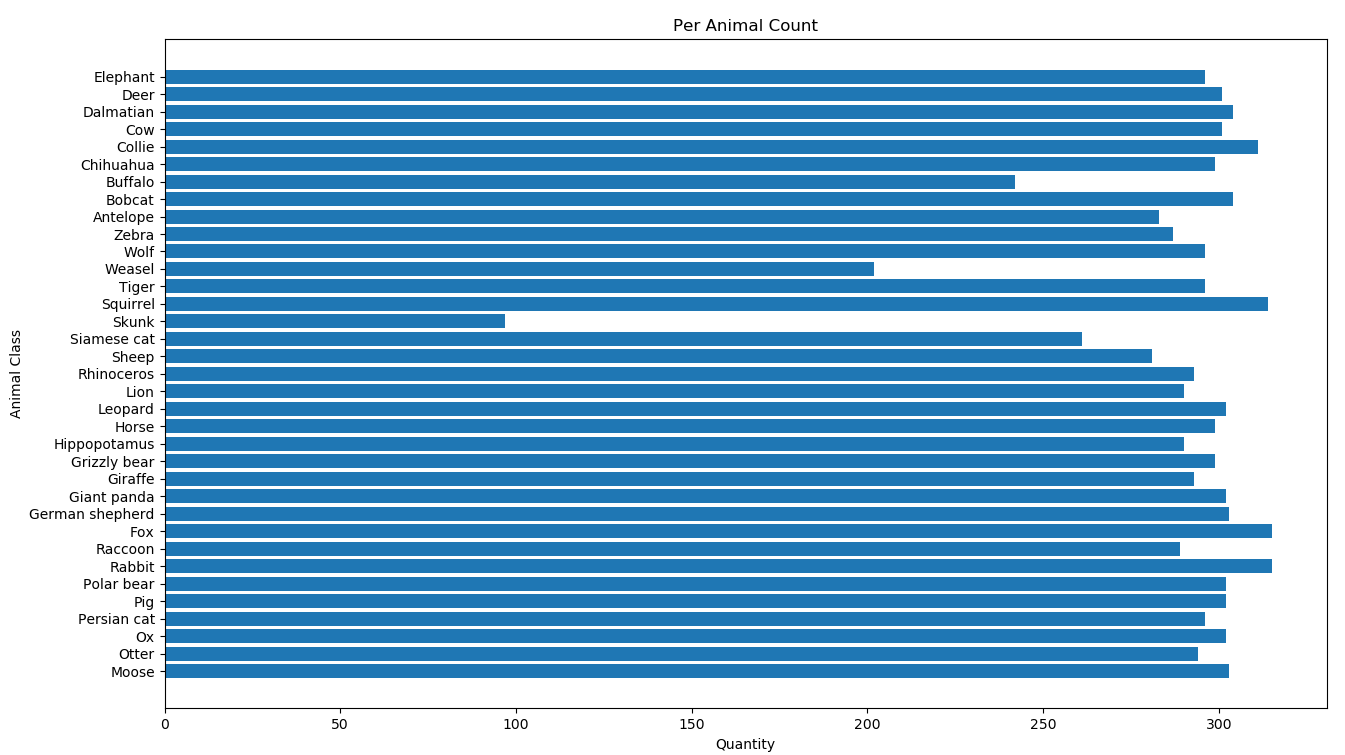}
\end{center}
  \caption{Number of images per animal class.}
\label{fig:per_animal}
\end{figure*}

After selecting the list of the keypoints, we selected images of 35 quadruped animal classes from the AwA dataset. For the current AwA Pose dataset, we only included the images where there is a single animal presenting. In the future, we plan to include multiple instance images to expand dataset. To annotate the keypoints and the bounding box of the animal in each image, we used the LabelMe \cite{labelme} tool. A sample annotation is shown in Figure \ref{fig:labelme}. As depicted, our keypoint set covers a lot of details of the animal body so they can be useful for various applications such as fine grained action recognition, part segmentation, and 3D reconstruction. Per image annotation took around 5 minutes. 
%All of the annotations are done by the first author of this paper.
%no need to emphasize this as all readers understand it. Once the paper is accepted, please add this information to the web that you will host the dataset and source code
We also tried to annotate roughly the same number of images for each animal class to make the dataset balanced. The number of annotated images per animal class can be seen in Figure \ref{fig:per_animal}. In total, we have annotated around 10k images with keypoint and bounding box annotations.

% \begin{figure*}
% \begin{center}
% \includegraphics[width=16cm]{images/heatmap.png}
% \end{center}
%   \caption{Heatmaps are generated for each of the keypoints by the HRNet model for a given animal image. The center of the heatmaps are considered as the predicted location of the keypoints. The figure is adapted from \cite{yin2021shetconv}}
% \label{fig:heatmap}
% \end{figure*}

\subsection{Model for Dataset Evaluation}
We have trained the HRNet \cite{xiao2018simple} model for detecting the keypoints from the quadruped animal images using our keypoint annotations. HRNet \cite{xiao2018simple} has shown superior performances for human keypoint detection problem thus should be suitable benchmarking model for our animal keypoint dataset. The HRNet \cite{xiao2018simple} model is a convolutional neural network model which can keep the 2D feature resolution high throughout the whole network. It has multiple scale streams where the topmost stream keeps the convolutional feature resolution same as the input image which helps to detect keypoints even when the animal body part is very small (e.g., eyes, nose tip, etc.) But large convolutional feature maps cannot learn global information due to the small convolution kernel size and large feature map size which are only suitable for learning local information (e.g., only eye or nose). To accurately detect each of the keypoints, it is important to have a global knowledge of all the keypoints locations. This will provide contextual information for detecting occluded and ambiguous keypoints. For example, if the `back left paw' keypoint is occluded by grass and the keypoint `back left knee' is clearly visible, the location of `back left knee' can help detect the keypoint `back left paw'. HRNet also downscale the feature size to various scales (e.g., 2x, 4x) and then uses skip-connections to upscale and feed them back to the original high resolution stream. This overall approach allows the network to detect keypoints efficiently from images even in difficult cases. The HRNet model generates heatmaps for each of the keypoints in our predefined keypoint dictionary. The center of the keypoints are considered as the predicted 2D locations of the keypoints.  As the images of our current dataset only contain a single animal, we did not use the annotated bounding boxes for training although they were used for evaluation.

\section{Experiments and Analysis}

To evaluate the HRNet for detecting keypoints on our quadruped animal keypoint dataset, we conducted several experiments for both seen and unseen animals. For each of the experiments, we examined the model performance for each individual keypoints.

\subsection{Training}

In this research, the HRNet-W48 version of the HRNet model was trained for our experiments. We used input image resolution 384x288 and output heatmap resolution 96x72. During the training time, we used data augmentations such as mirroring, rotating and scaling of both the training images and corresponding keypoint annotations. The minibatch size is 20 and the initial learning rate is 5e-4. Also, we used Adam optimizer for the training and trained the model for 120 epochs. For both the seen and unseen animal keypoint detection experiments, 5-fold cross validation was used. For the seen animal experiments, we randomly chose around 20 images per animal for testing. We selected roughly full body animal images for testing to use the bounding box as the scale for the animal for evaluation. For the unseen animal experiment, we randomly chose four animals for testing and the rest of the animals for training. Thus during training time, none of the animals in the test dataset are seen by the model. We conducted this experiment to see if a model can learn the generalized concept of the keypoints from our dataset. Experimentail results indicate that our dataset is particularly suitable for such applications as AwA Pose contains 35 different quadruped animals. 

\begin{table}[t]
\centering
\begin{tabular}{ccccccccc} 
\toprule
\multirow{2}{*}{keypoint} & \multicolumn{5}{c}{Seen Animals} & \multicolumn{3}{c}{Unseen Animals}  \\ 
\cline{2-9}
                          & @0.001 & @0.0005 & @0.0001 &  &  & @0.001 & @0.0005 & @0.0001          \\ 
\hline
Nose                      & 0.99   & 0.88    & 0.42    &  &  & 0.98   & 0.85    & 0.39             \\
Upper Jaw                & 0.99   & 0.88    & 0.45    &  &  & 0.96   & 0.83    & 0.4              \\
Lower Jaw                & 0.97   & 0.83    & 0.38    &  &  & 0.94   & 0.77    & 0.35             \\
Mouth End Right         & 0.97   & 0.79    & 0.34    &  &  & 0.94   & 0.71    & 0.27             \\
Mouth End Left          & 0.97   & 0.79    & 0.33    &  &  & 0.94   & 0.71    & 0.26             \\
Right Eye                & 0.99   & 0.91    & 0.55    &  &  & 0.98   & 0.88    & 0.51             \\
Right Ear Base            & 0.97   & 0.79    & 0.3     &  &  & 0.96   & 0.76    & 0.29             \\
Right Ear End             & 0.96   & 0.83    & 0.41    &  &  & 0.93   & 0.79    & 0.4              \\
Right Antler Base       & 0.97   & 0.74    & 0.3     &  &  & 0.93   & 0.73    & 0.22             \\
Right Antler End        & 0.88   & 0.68    & 0.37    &  &  & 0.73   & 0.57    & 0.23             \\
Left Eye                 & 0.99   & 0.92    & 0.54    &  &  & 0.98   & 0.89    & 0.53             \\
Left Ear Base             & 0.98   & 0.8     & 0.32    &  &  & 0.96   & 0.76    & 0.28             \\
Left Ear End              & 0.95   & 0.82    & 0.41    &  &  & 0.92   & 0.79    & 0.41             \\
Left Antler Base        & 0.97   & 0.79    & 0.34    &  &  & 0.96   & 0.8     & 0.29             \\
Left Antler End         & 0.89   & 0.75    & 0.47    &  &  & 0.85   & 0.67    & 0.26             \\
Neck Base                & 0.9    & 0.64    & 0.2     &  &  & 0.87   & 0.57    & 0.18             \\
Neck End                 & 0.84   & 0.54    & 0.16    &  &  & 0.78   & 0.49    & 0.15             \\
Throat Base              & 0.93   & 0.64    & 0.21    &  &  & 0.87   & 0.59    & 0.16             \\
Throat End               & 0.85   & 0.53    & 0.15    &  &  & 0.81   & 0.47    & 0.13             \\
Back Base                & 0.82   & 0.5     & 0.14    &  &  & 0.78   & 0.47    & 0.13             \\
Back End                 & 0.73   & 0.44    & 0.14    &  &  & 0.68   & 0.4     & 0.12             \\
Back Middle              & 0.73   & 0.45    & 0.14    &  &  & 0.68   & 0.4     & 0.12             \\
Tail Base                & 0.75   & 0.46    & 0.13    &  &  & 0.72   & 0.45    & 0.13             \\
Tail End                 & 0.74   & 0.51    & 0.17    &  &  & 0.7    & 0.49    & 0.17             \\
Front Left Thigh         & 0.8    & 0.44    & 0.12    &  &  & 0.79   & 0.42    & 0.11             \\
Front Left Knee         & 0.82   & 0.49    & 0.14    &  &  & 0.8    & 0.46    & 0.12             \\
Front Left Paw          & 0.88   & 0.62    & 0.19    &  &  & 0.87   & 0.61    & 0.18             \\
Front Right Thigh        & 0.81   & 0.44    & 0.13    &  &  & 0.8    & 0.42    & 0.11             \\
Front Right Paw         & 0.88   & 0.61    & 0.18    &  &  & 0.89   & 0.59    & 0.18             \\
Front Right Knee        & 0.82   & 0.5     & 0.12    &  &  & 0.83   & 0.47    & 0.12             \\
Back Left Knee          & 0.76   & 0.42    & 0.13    &  &  & 0.69   & 0.37    & 0.11             \\
Back Left Paw           & 0.83   & 0.55    & 0.17    &  &  & 0.82   & 0.52    & 0.17             \\
Back Left Thigh          & 0.73   & 0.38    & 0.11    &  &  & 0.68   & 0.32    & 0.09             \\
Back Right Thigh         & 0.74   & 0.37    & 0.11    &  &  & 0.69   & 0.35    & 0.1              \\
Back Right Paw          & 0.84   & 0.57    & 0.17    &  &  & 0.81   & 0.54    & 0.15             \\
Back Right Knee         & 0.76   & 0.42    & 0.12    &  &  & 0.73   & 0.37    & 0.1              \\
Belly Bottom             & 0.85   & 0.55    & 0.17    &  &  & 0.82   & 0.51    & 0.13             \\
Body Middle Right       & 0.69   & 0.33    & 0.1     &  &  & 0.66   & 0.3     & 0.09             \\
Body Middle Left        & 0.71   & 0.34    & 0.11    &  &  & 0.65   & 0.3     & 0.08             \\ 
\hline
Avg                       & 0.86   & 0.61    & 0.24    &  &  & 0.83   & 0.57    & 0.21             \\
\bottomrule
\end{tabular}

\caption{PCKB results for the seen and unseen animals for HRNet.}
\label{table:pckb}
\end{table}

\subsection{Evaluation Criteria}
To evaluate the model's performance on our dataset, we used the widely adopted metric, PCK (Percentage of Correct Key-points) \cite{andriluka20142d}. PCK considers a keypoint detection correct if the detection is within a given threshold. As the quadruped animals can vary widely in shapes and sizes, we used the bounding box annotations for the animals to calculate the threshold. We consider the keypoint detection correct, if the distance between the predicted keypoint location and the ground truth keypoint detection is within a certain value multiplied by the diagonal of the animal bounding box. This way, the evaluation is independent of the animal size. We named the evaluation criteria PCKB with B referring to the "bounding box". The calculation of PCKB is as follows: 

\begin{equation}
    PCKB =  \frac{1}{K} \sum_{i=1}^K \mathbb{I}[dist_{i} < SF * D]
\end{equation}

where $dist_i$ is the distance between the predicted and the ground truth locations of the $i$th keypoint; $D$ is the length of the diagonal of the bounding box; $SF$ is a scale factor for deciding the distance threshold in terms of $D$ (e.g., $PCKB@0.001$ is the PCKB value while $SD$ = 0.001); $K$ is the total number of visible keypoints; $\mathbb{I}$ is an indicator function which produces one if the condition is true, otherwise zero. PCKB is ranged between 0 to 1 and higher PCKB indicates higher detection accuracy.

\begin{figure*}
\begin{center}
\includegraphics[width=14cm]{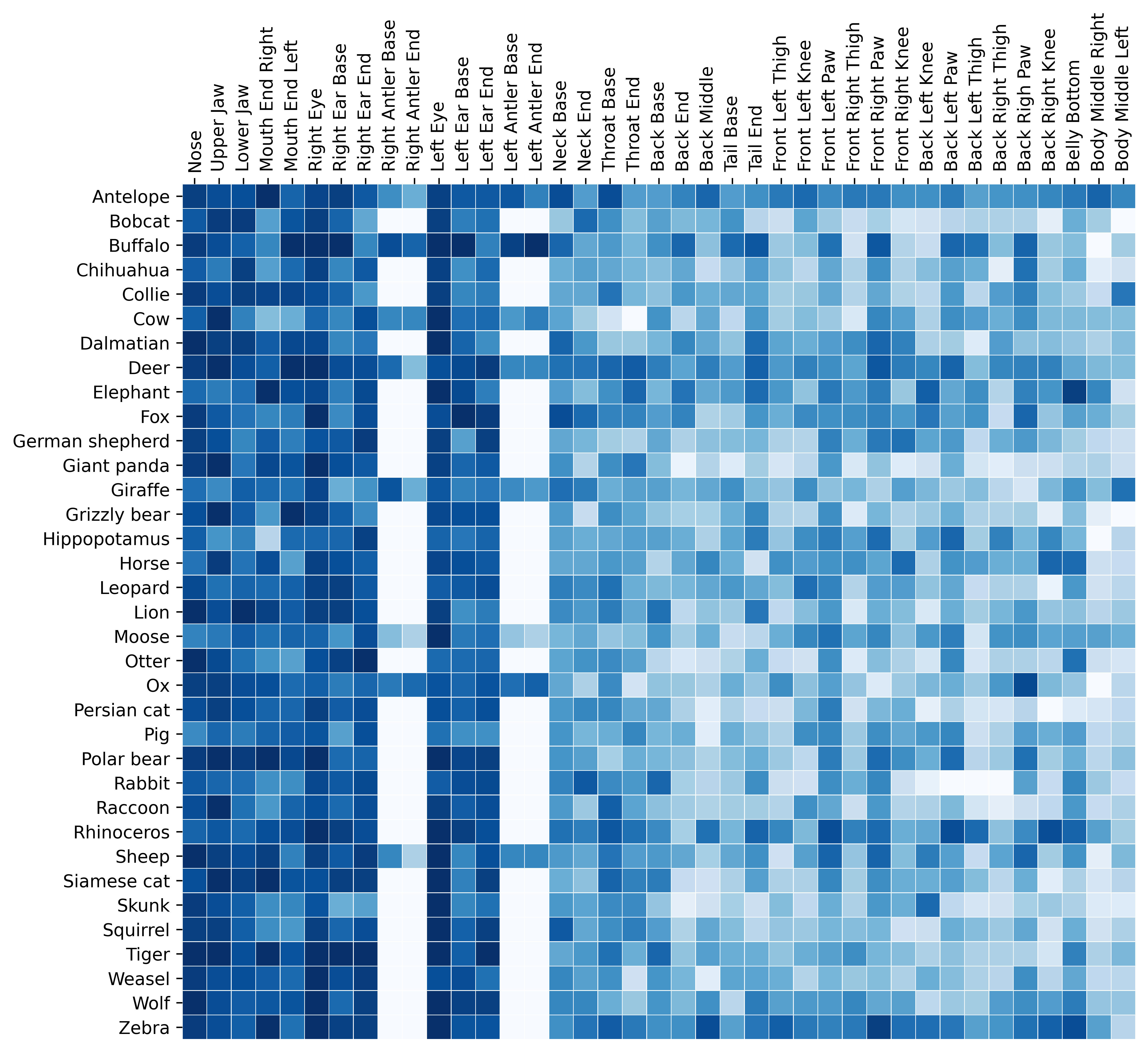}
\end{center}
  \caption{Per keypoint PCKB@0.0005 for each animal for the seem animal experiment. Dark blue indicates higher PCKB where lighter blue indicates lower PCKB. }
\label{fig:heatmap}
\end{figure*}

\begin{figure*}
\begin{center}
\includegraphics[width=14cm]{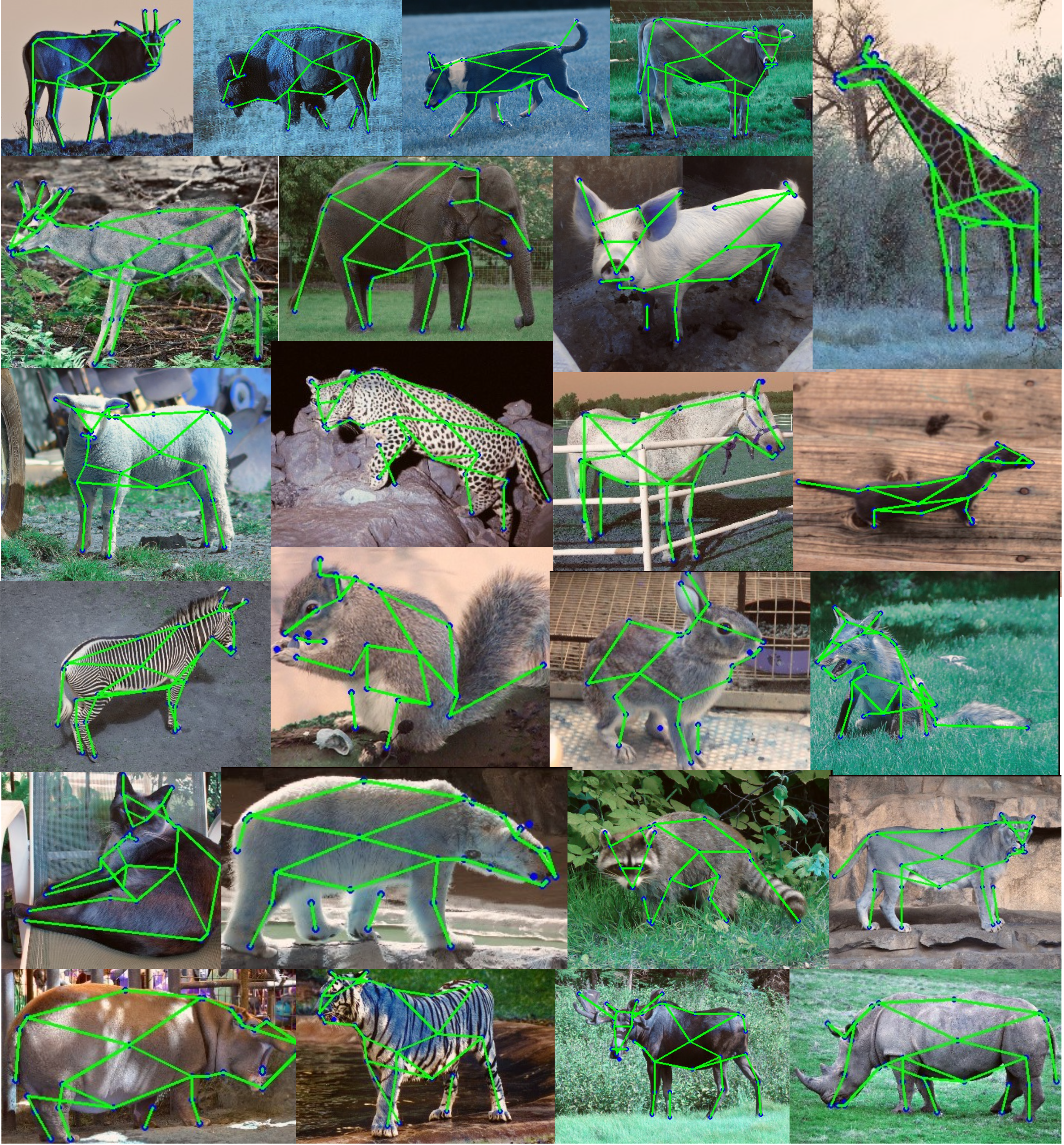}
\end{center}
  \caption{Qualitative results of the seen animal experiment. }
\label{fig:good}
\end{figure*}

\begin{figure*}
\begin{center}
\includegraphics[width=12cm]{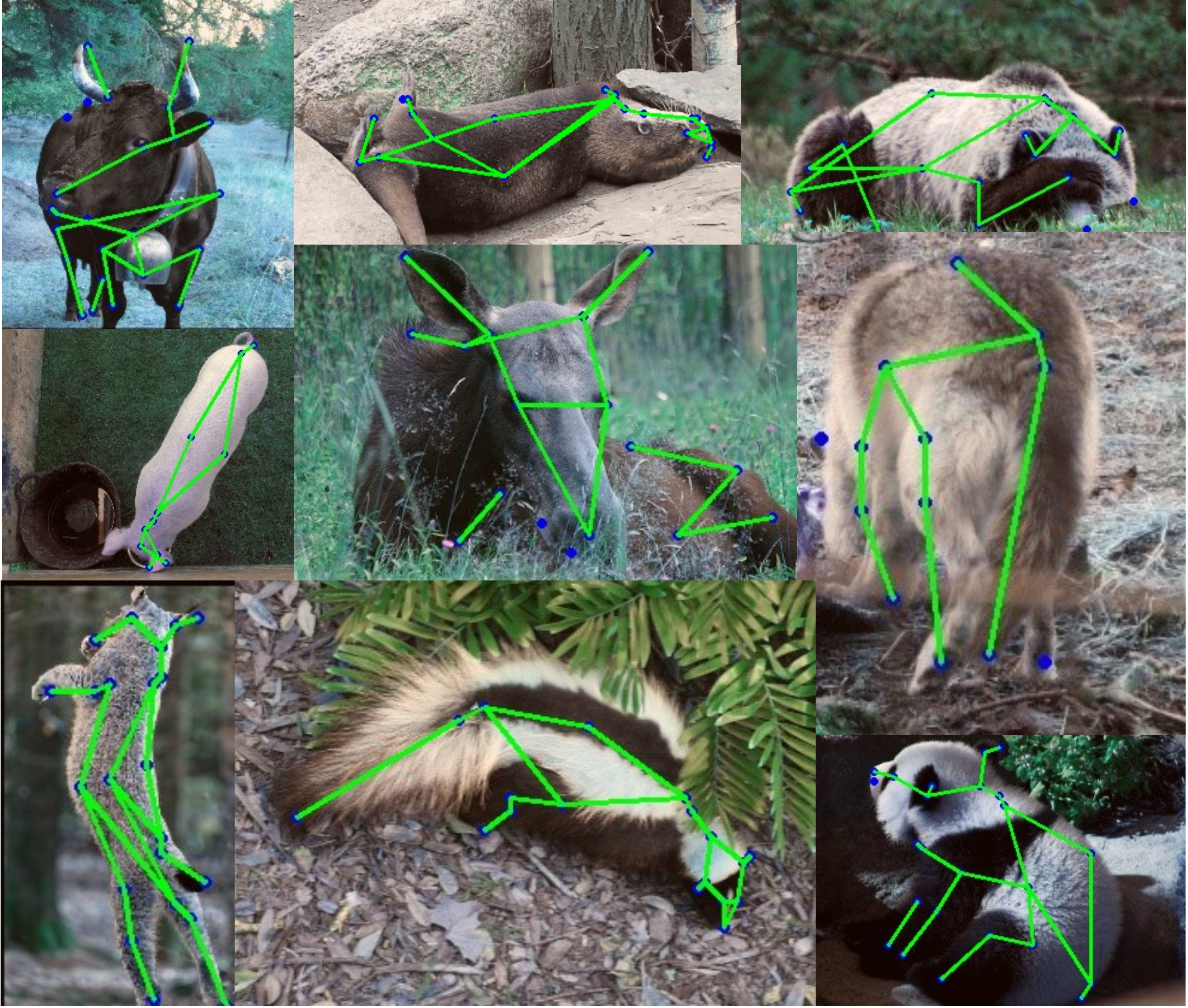}
\end{center}
  \caption{Failure cases for the seen experiment. }
\label{fig:bad}
\end{figure*}

\subsection{Results for Seen Animals}
Table \ref{table:pckb} presents the results of the seen animal experiments in which we used three different scale factors for the distance threshold calculation, 0.001, 0.0005, and 0.0001. The bigger the value, the more the tolerance is given while measuring correctness of the predicted keypoint location. Accordingly, we get higher PCKB values for larger multipliers and lower PCKB values for smaller multipliers. This also indicates the efficacy of our evaluation criteria, as lower distance threshold is making it stricter to pass a prediction as a success while higher distance threshold is making it comparatively easier. The results also show that the PCKB for the upper body keypoints such as the facial keypoints are much higher than the lower body keypoints. This is probably due to the complex and articulated poses presented by the lower body parts (e.g., legs), which is illustrated in Figure \ref{fig:heatmap}. A PCKB value was calculated for each pair of animal species and keypoint type. Daker blue indicates higher PCKB value and lighter blue indicates lower PCKB values. It can be clearly seen that the left part of the heatmap shows the higher values which represents the upper body keypoints and the right part of the heatmap shows lower PCKB values which represents mostly the lower body keypoints. Also, we can see that for most animals, the lowest body keypoints (i.e., paws) are well detected. But for small animals, such as rabbit, squirrel, and skunk, the lower body parts are more difficult to detect. On the other hand, some big animals such as giant panda can be difficult too as it's smaller body parts can be occluded by larger body parts. Moreover, for animals which often present complex poses, such as persian cat and siamese cat, it can be very difficult to detect the lower body parts. But overall, the model can localize the keypoints pretty well as shown in Figure \ref{fig:good}. Figure \ref{fig:bad} depicts some failure cases. In general, if the body parts are over occluded or the viewing angle is very different (e.g., top view), the model performance is poor.

\subsection{Results for Unseen Animals}
 We also conducted experiments for unseen animal keypoint detection where the model was trained on some animal species and then applied to a completely different set of animal species. The results are shown in Table \ref{table:pckb}. As we can see, the results for the zero-shot experiment are similar to the seen experiments. Although the PCKB values for the unseen animals are always lower than those of the seen animals, they are very close. This indicates that a model trained using our dataset can be generalized to unseen quadruped animals. This is due to the inclusion of diverse quadruped animals and dense keypoints in our dataset.
\section{Conclusion}
In this paper, we introduce a novel quadruped animal keypoint dataset---AwA Pose in which we annotated a large number of images with dense keypoints. Our dataset contains 35 different animal species and a total of 39 different keypoints. Both numbers are much higher than those of the existing datasets. Experimental results show that the dataset can be applied to both seen and unseen animal keypoint detection with solid performance. As detecting animal keypoints is significantly harder than human keypoint detection due to the animals' wide variance in sizes and shapes, we believe that our dataset will contribute to the computer vision community by reducing the gap between human and animal keypoint detection research. The dataset can be used as a test bed for novel keypoint detection models. It can also be used for more downstream tasks such as action recognition, part segmentation, 3D reconstruction, etc.

{\small
\bibliographystyle{ieee}
\bibliography{egbib}
}
\end{document}